\documentclass[11pt]{article}
\usepackage[utf8]{inputenc}
\usepackage[font=small,format=plain,labelfont=bf,up,textfont=normal,up]{caption}
\usepackage{natbib}
\usepackage{graphicx}
\usepackage[hang, flushmargin, bottom]{footmisc}
\usepackage[margin=21mm]{geometry}
\usepackage{url}
\usepackage{setspace}
\usepackage{adjustbox}
\usepackage{amsmath}
\usepackage{courier}
\usepackage{threeparttable}
\usepackage{booktabs}
\usepackage{subfig}
\usepackage{multirow}
\usepackage{csquotes}
\usepackage{hyperref}

\title{Machines Do See Color: A Guideline to Classify Different Forms of Racist Discourse in Large Corpora.}
\author{Diana Dávila Gordillo, Lake Forest College, ddavilagordillo@lakeforest.edu\\ Joan C. Timoneda, Purdue University, timoneda@purdue.edu \\ Sebasti\'an Vallejo Vera, Western University, sebastian.vallejo@uwo.ca (*Corresponding Author)\vspace{0.3cm}} %
\date{}

\begin{document}

\maketitle

\thispagestyle{empty}

\begin{abstract}
\noindent Current methods to identify and classify racist language in text rely on small-n qualitative approaches or large-n approaches focusing exclusively on overt forms of racist discourse. This article provides a step-by-step generalizable guideline to identify and classify different forms of racist discourse in large corpora. In our approach, we start by \textit{conceptualizing} racism and its different manifestations. We then \textit{contextualize} these racist manifestations to the time and place of interest, which allows researchers to \textit{identify} their discursive form. Finally, we \textit{apply} XLM-RoBERTa (XLM-R), a cross-lingual model for supervised text classification with a cutting-edge contextual understanding of text. We show that XLM-R and XLM-R-Racismo, our pretrained model, outperform other state-of-the-art approaches in classifying racism in large corpora. We illustrate our approach using a corpus of tweets relating to the Ecuadorian \textit{indígena} community between 2018 and 2021.
\end{abstract}

\vspace{1cm}
\begin{centering}
The full Appendix can be found at \url{https://svallejovera.github.io/files/Racism__Latin_America_and_NLP.pdf}.

\end{centering}
\vspace{1cm}

\pagebreak
\setcounter{page}{1}
\setstretch{2}

\section{Introduction}

A wide-reaching literature explores the political consequences of racial attitudes and racism \citep{sears2000racialized,tesler2013return,canizales2021latinos}, and the different forms racism takes \citep{bonilla2006racism,smith2020dynamics}. Researchers have studied how the overt nature of `old-fashioned racism' has been intertwined with new, more covert forms of racism \citep{desante2020fear}. Covert racism, similar to symbolic racism, implicit racism, racial resentment, or laissez-faire racism \citep{cramer2020understanding}, has served as an explanation for the continuing racial differences across social, economic, and political participation and achievements. This form of racism has shown resiliency despite the increasing social punishment and political repercussions of overt racism \citep{beauvais2022political}.


Many studies on racism and racial identity focus on political and economic inequalities and exclusions \citep{bonilla2006racism,cramer2020understanding} or the consequences of racist prejudices and attitudes \citep{harell2016race}. These often rely on measures of overt or `old-fashioned' racism \citep{tesler2013return}, or attitudinal measures of `racial prejudice' \citep{huddy2009assessing}. Yet, partly due to societal rejection, these overt forms of racism have become less common and more challenging to measure, particularly in text. At the same time, the perdurance of discriminatory practices highlights that racism continues to impact societal interactions. Our article contributes to the modest efforts in political science (see, for example, \cite{beauvais2022political}) that advocate for identifying and measuring more subtle forms of racism. Furthermore, our work addresses how racism is socialized or reproduced. While fundamental to our understanding of racism, current research does not address those aspects specifically. Racist behavior is learned, and the learning process is mainly discursive and through everyday interactions, (social) media, literature, political speeches, and other communicative events \citep{van2009racism}. Because language evolves alongside social norms, text is a rich source to identify and study the processes that activate, perpetuate, or stop racist behavior. 

Systematic identification of different forms of racist discourse has been limited to surveys \citep{rabinowitz2009white}, interviews \citep{marom2019under}, case studies \citep{bourabain2021everyday}, or anecdotal evidence \citep{levchak2018microaggressions}. Notably, while text-as-data has widened the scope of the social sciences \citep{grimmer2022text}, scholars have yet to exploit the wealth of information produced and reproduced in language, speeches, and social media to study racist content.\footnote{Cramer (\citeyear{cramer2020understanding}) encourages researchers to ``turn to methods that enable us to examine the language that people use while they are interacting with other people in order to better understand the role of racism'' (pg 161). By doing so, she argues, we can understand the socialization and transformation of race and racist messages in different spaces. We can achieve this by, for example, conducting ethnographies of online spaces or studying geographies' role in forming public opinion. Our proposed method facilitates large-n studies of both.} There is no methodological tool to systematically identify different forms of racist discourse, and racist discourse more broadly, in large corpora. Furthermore, the current computational tools used to detect a more general and less nuanced `hate speech' often lack the theoretical foundation to inform their models.\footnote{For example, \citep{zhang2019hate} develop tools to detect hate speech in Twitter data, yet there is no discussion on \textit{what} is hate speech, why they define hate speech the way they do, how their training set was built, or how it was validated.} 

To address the paucity of research on different forms of racism in discourse, this article provides a generalizable method to identify and categorize different manifestations of racism in large corpora. We present a guide that connects conceptual and contextual elements of racism and racist discourse, applies them to a coding scheme, and uses machine-learning techniques to categorize racist language in large corpora. Our process is highly flexible throughout its steps. It allows researchers to conceptualize racism in a way that aligns theoretically with their own work and research question; it can be applied to different geographical contexts and data sources; and while we encourage readers to use our proposed machine-learning approach, researchers can use their model of choice. 



\begin{figure}[!ht]
\centering
\includegraphics[width=10cm]{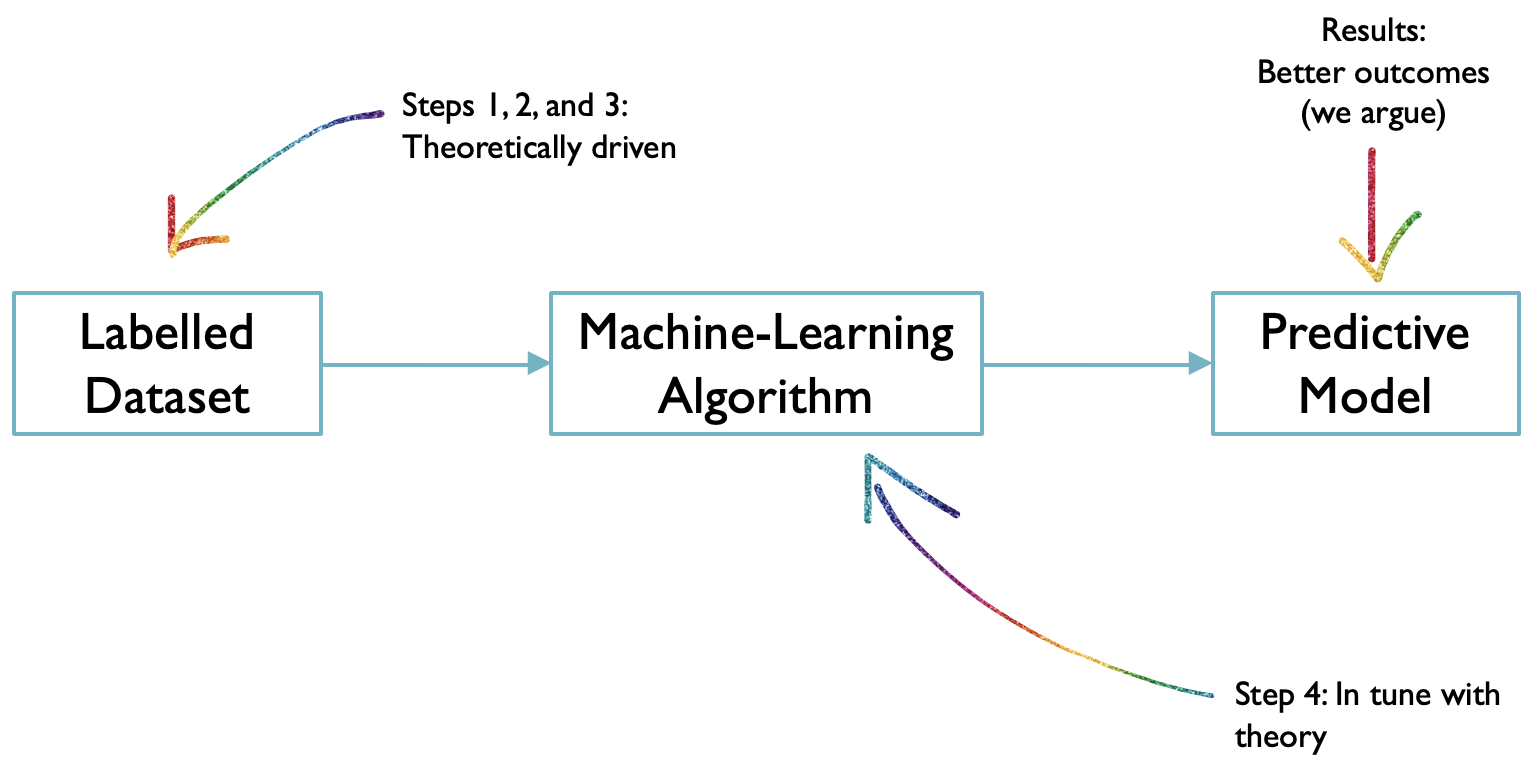}
\caption{We diagram the workflow presented in this article. Steps 1-3 produce the labelled dataset; step 4 trains the machine-learning model. The predictive model we ultimately obtain will lead to better outcomes.}
\label{fig:workflow}
\end{figure}

The step-by-step process described in the article requires researchers to 1) conceptualize racism, 2) contextualize racism and racist discourse, 3) identify the manifestations of racist language, and 4) apply a novel Natural Language Processing (NLP) technique to classify text. We start by 1) conceptualizing racism and how it is manifested. While our method is agnostic about the definition of racism chosen by the researcher, we stress the theoretical importance of this choice to the outcome. In this article, we broadly define racism as a system of racial domination \citep{bonilla2001white} and develop hypotheses based on this definition. 2) Regardless of the preferred definition, the historical development of racism, and its manifestations, are contextual to the place and time analyzed. For example, the colonial consequences of the Spanish legacy in Latin America developed a racist structure different from that imposed in British colonies. 3) Once identified the context, researchers must determine how racism and racist language manifest within that context. Coding schemes are developed using these contextual manifestations. We then use the coding scheme to label a training set, a sample of texts used to train our machine-learning model. Since racist language is contextual and is identified using contextual cues, we 4) train BERT-based machine learning models (i.e., models particularly adept at identifying context) to classify text accurately. Figure \ref{fig:workflow} presents a workflow diagram of the proposed approach.

The standard computational approaches employed to predict categories in text use labeled data sets to train machine learning models. Improvements to these models often focus on changes to the algorithm. We take a more holistic and theoretically-driven approach, paying attention to the information going into the model and how the model can accurately interpret this information.

To showcase the value of our process, we focus on covert and overt racist discourse in Ecuador. We propose a set of hypotheses that we evaluate using Twitter data between 2018 and 2021 relating to the Ecuadorian indígena community. We pay special attention to the  2019 protests. We describe the characteristics of racist content within that context, define the manifestation of different forms of racism in Ecuador, and use our prediction model to estimate user reaction to racist discourse in social media. While we look at Ecuador in this article, it is important to note that the process is agnostic to case selection and generalizable to other contexts and corpora. Researchers need only to replace Ecuador with their social context of choice. 

\section{Conceptualizing Racism: Overt and Covert Racism} 

As `old-fashioned racism' became socially punished in most contexts, new forms of racist discourse emerged (e.g., symbolic racism, covert racism, laissez-faire racism). Their social, political, and material consequences have been documented and understood as consequences of a historical process \citep{grosfoguel2016racism}, a structure \citep{bonilla2006racism}, or a behavior \citep{allport1958}. The mechanisms that reinforce (or challenge) racism are mediated through language and manifested through public discourse (e.g., political discourse). The first step to identifying racist speech, particularly as it evolves and becomes less explicit, requires researchers to conceptualize racism. Different understandings of racism, its origins, and its manifestations in social (i.e., political) spaces will yield different linguistic expressions. Thus, we start by answering the question: \textit{what is racism?}

\subsection{What is Racism?}

Scholars have conceptualized and understood racism differently, often as a complex and evolving notion. For \citet[][p.17]{fields2022racecraft}, racism refers to ``a social practice'' that routinely assigns people to categories that determine their social standing (writ large) on that basis. For the authors, race is a shorthand for racism, transforming an action into the characteristic of a subject. \citet[][p.1]{grosfoguel2016racism} develops Frantz Fanon and Boaventura de Sousa's theory of racism, understanding racism as a ``global hierarchy of superiority and inferiority'', drawing a line to determine who is considered human (and thus worthy of human rights), and who is considered sub-human (and thus unworthy of human rights). By defining racism in those terms, Grosfoguel recognizes that color is but one marker used by `the oppressor' to draw these lines across different geographies and social contexts.\footnote{Fields \citeyear{fields2001whiteness} makes a similar argument: ``From Peterloo to Santiago, Chile, to Kwangju, South Korea, to Tiananmen Square and the \textit{barrios} of San Salvador, humanity has learned again, and again that shared color and nationality set no automatic limit to oppression.'' However, they focus their study on the construction of race in the United States as an explanation for social phenomena (i.e., racecraft). Grosfoguel (2016), on the other hand, approaches racism through the lens of colonizer and colonized (i.e., the Boaventuran `abyssal line', or the Fanonian `line of the human').}

More familiar to the political science literature is Allport's (\citeyear{allport1958}) understanding of ``racism as prejudice,'' manifested as stereotypes towards the out-group and the innate inferiority of certain racial groups. From this concept, researchers theorized that racism was expressed more often as the \textit{perception} that certain racial groups do not adhere to \textit{American} values such as hard work, thrift, and patriotism \citep{kinder1981prejudice,kinder1996divided,sears2005over,hutchings2014racism}. These perceptions are socialized and reproduced through personal interactions, mass media, and religious institutions \citep{henry2002symbolic}, and operationalized as `racial resentment.'

In this article, we depart from the individualist framework of Allport and conceptualize racism as an ideology, explicitly connecting it to a system of racial domination \citep{bonilla2001white}. Ideologies create social hierarchies intrinsically connected to domination. Thus, racism is ``the product of racial domination projects (e.g., colonialism, slavery, labor migration, etc.), and once this form of social organization emerged in human history, it became embedded in societies'' \citep{bonilla2015structure}. These projects create \textit{races} out of people who were not so before: European peoples becoming White \citep{painter2010history}, African peoples becoming Black \citep{wright2004becoming}, and peoples of the Americas becoming \textit{Indios} \citep{bonfil1977concepto}. Racism is a combination of practices and behaviors that produce and reproduce racial structures and define social relations among newly created races with economic, political, and, overall, material consequences \citep{bonilla2015structure}.\footnote{Notice that these conceptualizations of racism are talking to one another, though not explicitly in academic dialogue. Thus, the coding rules we will produce might partially overlap, regardless of the definition.}

\subsection{Different Forms of Racism: Covert and Overt Racism} 

Racism is manifested in different forms. Common to most authors is the description of shifts in how societies engage with, tolerate, and express racism. For the ``racism as prejudice'' camp, old-fashioned racism--open bigotry stemming from a belief of the innate inferiority of a race--was replaced by ``racial anxiety and antagonism'' as minority groups gained political representation and rights \citep{mcconahay1986modern}. The prejudice towards minorities was framed as groups that are undeserving of their newfound prerogatives, given their lack of moral value. This individualist view of new racism has been described as \textit{symbolic racism} \citep{kinder1981prejudice,henry2002symbolic}, \textit{modern racism} \citep{mcconahay1986modern}, or \textit{racial resentment} \citep{kinder1996divided}.

An alternative view suggests that the changes in how racism is manifested derive from internal struggles to dismantle the racial structure. Yet, these struggles  are often limited by the possibilities of real alternatives \citep{bonilla2015structure}. For example, the dismantling of Jim Crow in the South after the Civil Rights Movement was partly a reaction to the instability created by riots and unrest and its effect on attracting capital \citep{bonilla2006racism}. The racial structure was maintained, as well as the material inequalities, even when the style of discrimination acquired new characteristics: the increasingly \textit{covert} nature of racial discourse and practices and the avoidance of direct racial terminology, among others \citep{bonilla2006racism}.

For our application, we are particularly interested in conceptualizing \textit{covert racism} as opposed to \textit{overt racism} (rather than, for instance, conceptualizing \textit{symbolic racism} as opposed to \textit{old-fashioned racism}). \citet{coates2008covert} defines covert racism as ``subtle and subversive institutional or societal practices, policies, and norms utilized to mask the structural racial apparatus.'' Crucially, as the racial structure is maintained, so are the material inequalities. Covert racism underpins racial profiling, disparities in access to information, disparities in access to resources and basic needs (i.e., housing), and disparities in prices paid for similar goods by different individuals. One of the main perils of covert racism is that it appears almost an innocent practice devoid of overtly racist content.

\subsection{Covert and Overt Racist Discourse} 

One way that societies reproduce racist practices is through language. Racist discourse not only reflects how people communicate these practices but also how they view ``power relations, normative frameworks, and the combination of symbolic and material reality'' \citep{herzog2022talking}. Covertly racist practices include covertly racist discourses. The discursive manifestations of covert racism are largely \textit{context dependent}. Racial stereotypes change from context to context as there are ``differential expectations for different racial groups'' \citep[][p.213]{coates2008covert}. The importance of context lies in the fact that covert racism perpetuates a particular racist social order. Its discursive form thus reflects this specific social order, which may need to be more evident to untrained (de-contextualized) eyes.\footnote{Of course, those who do not want to see such racist forms may also benefit from selective blindness.} Studying the discursive forms of covert racism requires deep knowledge of the context in which this discourse is employed.

Scholars have identified certain generalizable aspects of covert racist discourse, despite its contextual nature. We discuss five of these characteristics. First, it reproduces racial stereotypes with slightly sanitized terms \citep{shoshana2016language}—for example, the representations of Asian-Americans as `nerds' within the United States. The words are innocent enough to make racial elites able to participate in such practices, yet avoid potential social punishment \citep[][p.222]{coates2008covert}. Second, covert racism relies on racial codes, that is, ``words, phrases and/or ideas that may camouflage its true racist intent or purpose'' \citep{chin2015age,coates2008covert}. These codes are often presented (and represented) by local authority figures that use the language to perpetuate the practices of domination \citep{levchak2018microaggressions}. Third, covert racism is presented under a political correctness and `politeness' façade \citep{bonilla2002linguistics}. As \citet{bonilla2002linguistics} argues, this form of covert racism (defined by the author as `color-blind racism') is presented in a slippery and ambivalent form. In some cases, speakers preface racist statements with different forms of \textit{disclosure of non-racism} (``I am not a racist...'', ``I have [racial identity] friends...'') or clarify that the topic pertains to anything but race. Alternatively, speakers take both sides (the non-racist and racist sides) or aim to soften racist statements with diminutives \citep{bonilla2002linguistics}. The discursive ambivalence, and the semantic moves employed, can take different shapes under different contexts.

Fourth, covert racism discourse often takes the form of light talk --a hybrid between private and public sphere statements. These discursive forms presented as light talk hide racist content behind a veneer of playfulness \citep{levchak2018microaggressions} --for example, the use of Spanish words in English language statements with ambivalent meanings, e.g., ``\textit{adios}'' or ``\textit{hasta la vista}'' which could mean a formal farewell or a form of expulsion \citep{hill1995junk}. Finally, covert racism can take the form of ``microinvalidations.'' In this case, statements negate experiences or negate race entirely. Individuals are told not to be ``too sensitive'' or the speaker declares herself as ``unable to see race'' \citep{levchak2018microaggressions,shoshana2016language}.

\section{Contextualizing Racism and Racists Discourse}

The manifestations of racism are contextual. If symbolic racism is a product of prejudice and resentment towards other races, then the resentment will come from different historical developments: in the U.S. from affirmative action \citep{kinder1996divided}, in Europe from immigration \citep{blinder2013better}, and in Brazil from the myths surrounding their `racial democracy' \citep{telles2013understanding}. If covert racism is a product of changes in the racial structure, then the development of a racist structure and its manifestations are contextual to the place and time analyzed. For instance, the colonial consequences of the Spanish legacy in Latin America developed a racial/racist structure different from that imposed in British colonies \citep{katzew2005casta,rich1990race}.

\subsection{The Racial Structure in Ecuador} 

The racial structure in Ecuador is a product of the Spanish colonial domination project. The material and political inequalities begun with the \textit{hacienda} or \textit{huasipungo} system in the late 16th century, where Indians were expected to pay tribute to the landowners in exchange for the right to use a parcel of land in the hacienda's territory \citep{oberem1985sociedad}. After independence in 1822, the \textit{hacienda} system, as well as the \textit{indio} tribute, continued as large landholdings expanded. Only during the 1960s, with the land reform, oil boom, early industrialization, and expansion of the urban center did the \textit{hacienda} system withered out \citep{pallares2002peasant}. Nevertheless, the racial structure remained. Thus, the \textit{indígena} population started to organize and mobilize politically. This organization took the shape of  the \textit{indígena} uprisings of the nineties, massive mobilizations throughout the 21st century, and the foundation and institutionalization of a political party.  The indigenous population  demanded and eventually conquered important political and social victories \citep[see][]{pallares2002peasant}.\footnote{The indígena population has organized around the ``indígena" identity. Note, however, that different and sometimes conflicting groups form it.} Yet, rejection of their demands from the mestizo and blanco-mestizo population has been constant, as have racist attacks and material inequalities.\footnote{We provide a more nuanced historical look at the racial structure in Ecuador in Appendix A.}

The state structures in Ecuador have perpetuated unequal systems dominated by a blanco-mestizo dominant ideology \citep{roitman2017mestizo}. Research on race and ethnicity in Ecuador shows that \textit{mestizaje} has created a `whitening' process. Mestizaje, and also most people who identify as mestizos, downplays discrimination towards indígenas and other marginal communities while still engaging in \textit{covert} racist practices \citep{beck2011que}. Thus, in Ecuador, discursive racism is usually framed as an acceptance or tolerance of the out-group (e.g., indígena) by creating strict demarcations between the self and the `other.' Unsurprisingly, covertly racist language is normalized while having different forms. Individuals and society at large cover the racist elements of their actions and discourses through rhetorical means \citep{traverso2005discursos}, which leads to variation in the manifestations of racist discourse. Everyday patterns of behavior and speech, as well as the organization of the state, are constructed in a way that `\textit{indios}' and `indígenas' are the subjects and objects of structural discrimination.

Overall, racism in Ecuador is a ``system of ethnic-racial dominance" historically rooted in European colonialism \citep{van2009racism}. It is directed, in great part, towards the indígena population. The indígena population has been marginalized by a state that has done little to support these communities or grant them equal access to political spaces. Despite their political activism and mobilization, indígenas and the indígena community remain the main, yet not the only, target of the national racist ideology. Thus, in a country such as Ecuador, all platforms and institutions, classes, and contexts, reproduce a blanco-mestizo racist ideology in often subtle and normalized patterns that many times the user is not aware of or would not consider racist \citep{roitman2017mestizo}. This makes covert racism pervasive, which opens an ideal opportunity to test our method.

To showcase the usefulness of our method, we propose two exploratory hypotheses derived from our understanding of racism and the racial structure in Ecuador. Our first hypothesis focuses on the characteristics of the different types of racist discourse and their effect on behavior. Covert racism has replaced the overt manifestations of the racist ideology, partly as an answer to demands to dismantle the racial structure without actually changing it \citep{bonilla2015structure}. Furthermore, unlike covert racism, overt racism is socially punished (i.e., a socially costly behavior). Thus, our first hypothesis expects that

$H1_a$: Covert forms of racist discourse will be more prevalent than overt forms of racist discourse.

Given the costs associated with overtly racist behavior, an extension of our first hypothesis is that

$H1_b$: public figures (i.e., elites) are less likely to produce overtly racist discourse than non-public figures (i.e., the masses).

Our second hypothesis explores the prevalence of different forms of racist discourse in Ecuador. The indígena community in Ecuador has gained political space through pressures from indigenous organizations and protests. Part of their repertoire is the \textit{levantamiento indígena}. This is a strike that mobilizes different indígena organizations from their communities towards the capital. These often challenge blanco-mestizo power by disrupting political and economic life \citep{guerrero1997poblaciones,Bonilla_Mancero_2020}. The political psychology literature suggests that, in political contexts, threats to the identity of the in-party or to the status of the in-party \citep{mason2016cross} are drivers of out-group hate and anger. Manifestations of covert and overt racism will fluctuate according to the threat levels towards the blanco-mestizo population. Thus, we expect that:

$H2_a$: The prevalence of covert and overt racist discourse will increase when the indígena community challenges the status of the white-mestizo population.

As previously discussed, covert racism is normalized in Ecuador. More overt threats to the status of the blanco-mestizo population (e.g., levantamientos) should produce more overt manifestations of racism. Thus, we also expect that:

$H2_b$: The prevalence of overt racist discourse will increase at a faster rate than covert racist discourse when the indígena community challenges the status of the white-mestizo population.

To test these hypotheses, we use Twitter data from 2018 to 2021 relating to the Ecuadorian indígena community, focusing mainly on the 2019 protests. In addition to the abundance of text, this corpus includes information (i.e., tweets) from public figures and less-prominent users during the political turmoil that highlighted the historical confrontation between the indígena and the blanco-mestizo community \citep{vallejo2023rage}.

Note that researchers using other concepts of racism, other forms of racism, and/or other study cases need only to adapt the current process to their own necessities. Their research questions will also yield different hypotheses. For example, suppose we want to explore how anti-discrimination laws change racist language in political discourse. In that case, somebody studying symbolic racism in Brazil, where hate speech is criminally punished, might focus on how racial priming is manifested in Brazilian politics. 

\section{Identifying and Coding Racist Language}

We now turn to identifying and coding the different manifestations of racist discourse. In this section, we describe how we identify racist language in Ecuador, build a codebook, and provide a guideline for generalizing this codebook. Additionally, we describe the data source to be labeled and present descriptive statistics of the final labels and inter-coder reliability. In Appendix C, we provide details on training coders, lessons learned, and suggestions on tackling drawbacks found in the process. 

\subsection{Manifestations of Racism in Ecuador}

In order to identify the various manifestations of racism in Ecuador, we surveyed the extant literature on racist practices and discursive forms specific to our case. We started by asking, how is the racial structure manifested in modern-day Ecuador? As discussed above, recently, researchers have focused on the organization of the indigenous population. This work builds on prior research on the social and political standing of the indigenous population in Ecuador. These works highlight that a) diverse forms of racism, racial stereotyping, and racial discrimination are entrenched in Ecuadorian society \citep{beck2011que}, and b) that the intersection of ethnicity, race, and class has created spaces where race is used both as a mobilizer and as a social marker \citep{whitten2003symbolic}. Afterward, we returned to more recent work focusing solely on racist practices and discourse to ensure we covered the breadth and depth of research on the  topic. Our work was to systematize and bring together their insights applying our definitions of covert and overt racism. 

For our case, we define overtly racist language as any discourse that falls within one of the following categories: 1) discourse that explicitly includes derogatory terms or phrases that have been historically used to characterize the indígena population (either as individuals or as a community) as the lesser and dominated group in Ecuadorian society; 2) insults directed towards members of the indígena community that explicitly include their identity; 3) aggressive or denigratory language that includes the word `indio'; 4) racialized phrases or idioms; and 5) violence or incitement to violence towards members of the indígena community. Additionally, we define covert racist language as any discourse that describes the actions or character of the indígena population (either as individuals or as a community) by reproducing the idea of them as the lesser and dominated group in Ecuadorian society through masked, sanitized, or de-racialized language. The codebook used is available in Appendix C, where we provide details and examples of each category.  

Some sources referenced \textit{did not} use our terminology (i.e., covert and overt racism) to describe the manifestations of racism. Nonetheless, we parsed the manifestations of racism studied by these scholars following our definitions of covert and overt racism. 
Overall, we identify five general manifestations of covert racism: 1) \textit{no-difference racism} \citep{bonilla2006racism} or negation of identity \citep{canessa2007indigenous}; 2) \textit{attacks on the capabilities of the indigenous population} \citep{roitman2017mestizo}; 3) \textit{infantilization} of the indígena population; 4) \textit{hygienic racism} or deeming the indígena population unclean (metaphorically or literally) \citep{colloredo1998dirty}; and 5) \textit{ventriloquism} \citep{guerrero1997construction}. We provide extended descriptions for each manifestation in Appendix B. Since we are not interested in independently coding these different manifestations but in identifying covert racist language more broadly, we are not concerned with slight overlaps in the definitions.\footnote{For our example, overlaps avoid gaps in the general identification of covert and overt racism. However, researchers interested in coding each manifestation separately should make the definitions of the different concepts distinct and exclusive.} In Appendix B, we also include examples of covert and overt racist language from our corpus.

\subsection{Data} 

To build our training set and test our hypotheses, we use Twitter data covering indígena-related discourse in Ecuador. The main corpus of tweets was collected during the indigenous protests in Ecuador between October 1st and October 30th, 2019.\footnote{We collected the data by connecting \textit{rtweet} to Twitter's backward search application programming interface (API). See \citet{timoneda2018world} for a description of how the Twitter API works. Authors should also consider using Google Trends to obtain data on hard to reach populations \citep{chykina2018using, timoneda2022spikes, timoneda2021will}. We used the following terms in the search: \textit{paro} and \textit{ecuador}.} The corpus includes 2,020,487 posts (168,933 unique posts) by 91,458 unique Twitter users. A second corpus of tweets includes posts mentioning the indígena community in Ecuador between 2018 and 2021.\footnote{We used the following term in the search: \textit{conaie}, \textit{indio ecua}, \textit{protesta ecua}, \textit{indígena ecua}, \textit{mestizo ecua}.} The corpus includes 1,497,369 posts (154,630 unique posts) by 66,574 unique Twitter users. This second corpus allows us to test generalizability and control for inter-temporal losses in accuracy.

We used 3,724 tweets to create the training set. We sampled the corpus using three different approaches to have a more balanced training set. The first third of the training set was randomly produced from the entirety of the corpus. Another third of the training set was randomly produced from tweets that included the word `indígena' or `indio.'\footnote{An alternative, more charged term to refer to indígenas is ``indios.'' ``Indio'' (Indian) is often used by the blanco-mestizo population as a derogatory identifier. Despite the long history of the indígena community reclaiming the term, it is still used to signal a racial attack. Note, however, that using ``indio'' does not automatically reflect racism.} The final third of the training set was randomly produced from tweets that contained linguistic markers (i.e., words or phrases) that are commonly associated with the indígena community, including those that directly pertain to the indigenous identity (see Table B.1). In Table \ref{tab:examracism}, we provide examples from the data for each form of racist discourse previously described. 

\begin{table}[!ht] \centering
\scalebox{0.9}{\begin{tabular}{p{5cm} p{12cm}} 
\\ \hline \\
\multicolumn{2}{c}{\textbf{Covert Racism}} \\ \hline \\
\multicolumn{1}{l}{\textbf{Type}} & \textbf{Tweet (example)} \\ \hline \\
No-difference racism & ``This MESTIZO [sic] just like all ecuadorians is called CARLOS PEREZ, que disguises as indígena y make people call himYaku, Jah!!'' \\ 
Attacks on the capabilities & ``The [CONAIE] is a threat to the progress of the indigena population. Instead of destroying, focus on building a better country. Focus on educating and getting them out of their ignorance.'' \\ 
Infantilization & ``It is enough for Rafael Correa @MashiRafael to tweet his messages. It says a lot about you and the movement your lead @jaimevargasnae. They use the indigenas only to benefit the leaders, that is to say you!'' \\ 
Hygienic racism	 & ``we will play carnaval* with the indígenas, they fear water.'' \\ 
Ventriloquism & ``At this point this is not a strike for the economic policy. This is to destabilize, they are affecting the infrastructure of the State. The indigenas are stooges, they will not negotiate, they want to overthrow @Lenin.'' \\ \hline \\
\multicolumn{2}{c}{\textbf{Overt Racism}} \\  \hline
\multicolumn{1}{l}{\textbf{Type}} & \textbf{Tweet (example)} \\ \hline \\
Ethnic slurs & ``¡¡¡Damned \textit{Longo}!!! ¡you will not come back!'' \\ 
Attacks explicitly mentioning the ethnic identity & ``This is one is stupid even when they fix their stupidity. Who told him it is a country inside of another country? ¡Stupid indio!'' \\  \hline
\end{tabular}}
\caption{Examples of racism in Ecuadorian Twitter. All tweets are translated from Spanish. See Table B.2 for the original text.}
\label{tab:examracism}
\end{table}

\subsection{Codebook, Labeling, and Inter-Coder Reliability} 

The complete codebook from our example used to train coders is available in Appendix C. The codebook structure took the following form: 1) a general definition of covert and overt racism, 2) rules, explanation of rules, and examples of overt racism, 3) rules, explanation of rules, and examples of covert racism, 4) advice for coders to handle unclear/ambiguous text. To generalize this codebook, we encourage researchers and practitioners to maintain a similar structure: a general definition of the different forms of racism to be coded, followed by the various manifestations of those forms that coders might encounter in the training set, and advice on how to handle unclear or ambiguous text. The codebook structure allows researchers to label different forms of racism while providing important details of each category that  help coders have a more homogeneous understanding of each element. The final form of our codebook resulted from numerous training and discussion sessions with coders. One of the greatest difficulties we found in this process was labeling discourse that is, by nature, covert (e.g., covert racism, symbolic racism, laissez-faire racism, etc.). Thus, in Appendix C, we add details on training coders, lessons learned, and suggestions on tackling drawbacks found in the process. 

For this particular exercise, we followed \citet{schreier2012qualitative} in the coding process. We trained two hired coders across three review rounds to ensure consistency and performance. The initial review round (500 tweets) allowed us to explain discrepancies and identify cases we did not initially consider. After a second review round (500 tweets), coders independently coded 2750 tweets (Cohen score = 0.9 or strong agreement). In Appendix C, we provide details on revisions to the codebook to avoid ambiguities, common questions from the coders, and lessons learned throughout the process that can be generalized to other cases and can be helpful to researchers. 

To harmonize our training set when encountering coding discrepancies, we code as covertly racist or overtly racist if both coders agree. 
In Table \ref{tab:sumstat}, we present the distribution of categories in our training data. Notice that the data is skewed in favor of non-racist tweets, which is expected. Racism, while not uncommon, is not the main lexical form found in social media, as it is, we argue, a costly social behavior that can have real-life consequences. It is also monitored, flagged, and eliminated from most social media platforms.\footnote{However, Twitter's automatic detection of hate speech in Spanish is rather lax.}

\begin{table}[!t]
\centering \caption{Training Data by Type of Tweet \label{tab:sumstat}}
\begin{tabular}{l l l}\hline
\multicolumn{1}{c}{\textbf{Type}} & \textbf{N}
 & \textbf{\%}\\ \hline
Non-racist & 2,187 & 58.7  \\ 
Covert     & 1,035  & 27.8  \\ 
Overt      & 501  & 13.5  \\ \hline
Total      & 3,723 & 100 \\ \hline
\end{tabular}
\end{table}

\section{Detecting Racist Language Using Machine-Learning}

We use the labeled data as a training set to train a machine-learning model that can predict racist content in large corpora. To identify racist discourse, we often use context cues, as certain expressions or descriptions are racist depending only in the context in which they are used. We thus read a text as a whole rather than focus on the discrete meaning of each independent word. Therefore, we need a machine-learning technique that is particularly adept at understanding \textit{context} if we are to build an accurate classifier for covert and overt racist text.\footnote{Issues of access, resources, and technical knowledge can limit the options for researchers. While we encourage researchers to use our model of choice (for the reasons we describe in this section), the method proposed is agnostic to the ultimate model chosen. This also means that, as new models are developed, these can be adopted into the process described in the article.} This section introduces Transformers, a context-understanding machine-learning architecture used for text classification. We also explain how to exploit the full functionality of XLM-RoBERTa, the Transformers-based model we train to detect racism in text. We also briefly introduce the most common natural language processing (NLP) approaches in the social sciences, which we use as baselines to compare to our main model. Throughout this section, we use the Twitter data corpus from the Ecuadorian protest in 2019 described in the previous section.

\subsection{Pretrained Contextual and Non-Contextual Embeddings}

One of the first revolutionary advances in NLP for supervised text classification was word embeddings \citep{mikolov2017advances}. Word embeddings are mathematical representations of words, which can be used in machine learning models to classify text, among other applications. A commonly used English-language model is Word2Vec, which is trained on a large set of news text and contains 3 million word embeddings, or mathematical representations for 3 million different words (each embedding is a numeric vector of size 1 x 300).\footnote{Other popular word embeddings are GloVe and fastText. We use GloVe, which are considered state of the art, for the baseline models.}

However, the problem with these approaches is that word vectors are static, meaning each word has a corresponding fixed mathematical vector after training. For instance, the numeric vector for the word `bear' is the same in `grizzly bear,' `teddy bear,' `bear fruit,' or `bear a loss.' This is where the Transformers deep learning architecture innovates: it can dynamically capture the different meanings of words based on \textit{context}. In the example above, a Transformers-based model would produce four different word embeddings for `bear,' one for each word's specific use. Transformers generate dynamic word embeddings that change for every word depending on the context \citep[see][]{tunstall2021}.The dynamic nature of these embeddings allows for greater contextual understanding as the context largely determines the word embedding itself,
which makes Transformers models especially well-suited for an application such as detecting covert and overt racism in text.



\subsection{Bidirectional Encoder Representation from Transformers}

BERT stands for Bidirectional Encoder Representations from \textit{Transformers}. Transformers are large neural networks that produce representations (i.e., embeddings) of text input through the self-attention mechanism \citep{vaswani2017attention}. Self-attention relates each word to all other words in the sentence, which makes BERT and similar models \textit{bidirectional} in nature. Unlike other sequential and unidirectional methods such as Word2Vec, Transformers-based models can process tokens in a sentence \textit{all at once}. Transformer-based models are trained using an encoder-decoder architecture where input passes through a set of encoder layers that yield a \textit{representation} of the original input. We then use this representation to fine-tune a model or decode it to produce an output such as a translation into another language or a classification task \citep{ravichandiran2021getting}.

Google AI's BERT and Facebook AI's RoBERTa and XLM-RoBERTa are \textit{encoders} of a Transformer model. Facebook AI's innovative cross-lingual model XLM-RoBERTa, for example, consists of 24 encoder layers and 16 attention heads in its largest configuration.\footnote{Given the high computational requirements of BERT and RoBERTa models (as well as their cross-lingual variants mBERT and XLM-R), they both have `large' and `small' versions of their models. Large models are pre-trained on more data and contain more tokens, which tends to improve accuracy.} For this configuration, each layer's output is a vector with a length of 1024 \citep{ravichandiran2021getting}. Encoder layers are each step through which an encoder model converts input into a more accurate representation of that input. Attention heads are components of a layer that apply the Transformers architecture mentioned above by relating all words to other words. After each encoder layer, the representation becomes increasingly accurate, with each token incorporating more information from all other tokens to generate greater contextual understanding.

Using this set-up, BERT models and variants such as RoBERTa are pre-trained via large amounts of pre-existing text. The creators of BERT used 11,038 books from the Toronto BookCorpus and all of English Wikipedia to train BERT, for a total of 16GB of text. The cross-lingual RoBERTa, XLM-RoBERTa, was trained using filtered data from Common Crawl, and it contains 250,002 unique vocabulary elements \citep{ravichandiran2021getting, tunstall2021}. For words that do not directly match a token, XLM-R adds different word chunks (tokens) together and extracts a combined representation. For instance, the word `training' would be tokenized as ``train, \#\#ing'', with the double hashtag indicating that `ing' is a subword token that follows the token `train'. Subword tokenization strategies, or out-of-vocabulary strategies, have proven quite accurate at handling unknown words.\footnote{BERT-based models use different methods to generate tokens for subwords. BERT uses byte-pair encoding, which creates character groups according to their frequency in the training data. RoBERTa uses Byte-level Pair Encoding, which is helpful in multilingual contexts.} However, if a word is important enough to a researcher's specific application, these Transformer-based models can be further pretrained. In this article, we detail how this pretraining process works and leverage the flexibility of XLM-RoBERTa to improve its recognition of racist text in Spanish. Lastly, note that RoBERTa and XLM-RoBERTa models learn through Masked Language Modeling (MLM).\footnote{See Appendix D for a detailed explanation.}

With RoBERTa and XLM-R models, we can \textit{fine-tune} the model or \textit{further pretrain} the model. Most of the time, researchers fine-tune RoBERTa and XLM-RoBERTa models for a specific classification task. Fine-tuning means that we take a pre-trained RoBERTa and XLM-RoBERTa model and modify the last layer of the Transformers network, adjusting its weights so we can best predict the categories specific to our task. For instance, in this article, we fine-tune multiple models to classify text in Spanish from Twitter as either `covert' racism, `overt' racism, or `no racism.' We can then save this fine-tuned model and apply it to new, unseen tweets.

On the other hand, further \textit{pretraining} a RoBERTa or XLM-R model means that we add new word tokens to the original model, provide new text containing those tokens in context, and re-train the model. This approach leverages RoBERTa and XLM-R's original understanding of text and adds a new level of comprehension for the task at hand. After pretraining, we save the model and apply it to text in the same way we apply the original RoBERTa and XLM-R models. We explain this process in detail in the following subsection.


\subsection{Training an XLM-RoBERTa Model to Detect Racism in Text}

In all supervised machine learning models, there are three important parts to consider: the input text or training data, the model to train or fine-tune, and the testing and validation of the results. For the first step, we have provided readers with a detailed account on how to build a theoretically-grounded and rigorous training set. For the next two steps, we (1) pretrain and fine-tune an XLM-RoBERTa model built specifically to detect racism in Ecuador, and (2) apply 10-times repeated 10-fold cross-validation, reporting out-of-sample performance averages for model testing. Below we describe these steps in detail and present the main results of the article.


\subsubsection{Pretraining the XLM-RoBERTa}

Our model of choice is XLM-RoBERTa (XLM-R), Facebook AI's innovative cross-lingual model pre-trained on large corpora on Common Crawl, Wikipedia, and Open WebText data.\footnote{We use a version of XLM-R that is further pretrained in Spanish named entity recognition. The model's name is `xlm-roberta-large-finetuned-conll02-spanish' and can be found at \url{huggingface.io}.} XLM-R has been shown to produce high accuracy scores in multilingual applications \citep{conneau2019unsupervised}.

Timoneda and Vallejo Vera (2024) show that further pretraining BERT-based models significantly increases performance. For instance, by further pretraining XLM-R, we can generate a more accurate classification model to detect complex phenomena in languages other than English and for corpora with distinct vocabulary and linguistic constructions (for our example, Twitter data from the Ecuadorian protests in 2019). We pretrain our models in four steps: 1) we add new tokens to the original XLM-R tokenizer that reflect different expressions of racism in Ecuador; 2) we assign the mean representation of similar words to the newly added tokens; 3) we feed a new corpus containing the added tokens to the model and train it again to improve the representation of those words; and 4) we save the new model and apply it to our classification task (fine-tuning) in the same way we would apply the original.


Following the steps detailed above, to pre-train our own XLM-R model, we first add 20 tokens to the XLM-R tokenizer for a total of 250,022 tokens. We produced the list of 20 tokens based on our knowledge of the Ecuadorian context and what we learned while labeling our training data. Specifically, we found a series of terms that strongly signaled overt and covert forms of racism and shorthand expressions used, in part, to avoid being flagged as inappropriate content by Twitter (e.g., instead of ``hijo de p*ta'', users would write ``hdp''). These words were either not in the pre-trained vocabulary or appeared in an unrelated context (e.g., ``longo'' is a derogatory term in Ecuador, yet it only appears in the pre-trained data as the Portuguese word for ``long''). Thus, we added tokens such as \textit{longo}, \textit{guangudo}, \textit{poncho dorado}, and \textit{cholo}, among others, to the tokenizer.\footnote{See Appendix E for a complete list.} 

Tokens added to the vocabulary are given, by default, a random embedding, which we then replace with the mean embedding for similar preexisting derogatory terms.\footnote{Similar terms are synonyms or words used for a similar purpose. For example, when adding insults to the vocabulary, we give them the mean embedding of other insults. Thus, we provide to the added token `mamaverga' the average embeddings of `fucker', `idiot', `dick', and `stupid', all insults used in similar contexts. For a complete list of added tokens and their initial embeddings, see Appendix E.} Doing this imbues these tokens with initial substantive meaning, which helps the model recognize their contextual relationship to other tokens during training.

Afterwards, we use a new unlabelled set of over 1.7 million tweets to pretrain the model on new text.\footnote{For the unlabelled set we downloaded tweets that included the following terms: \textit{indio}, \textit{indígena}, \textit{longo}, \textit{guangudo}, and \textit{yunda}. The set covers tweets produced between January 2018 and December 2020.} The new corpus includes the added tokens. The new text allows the model to see how and where these words are used, leading to better understanding of the added tokens in context. The unlabelled Twitter corpus also provides additional information on the unique linguistic construction of tweets. During this additional pretraining step, we use the new data and retrain the model with it. In this process, the embeddings for the new tokens will change, making them a more accurate representation of their actual meaning. Furthermore, the embeddings of tokens that rarely appear in the original data but that are used often in the new unlabelled corpus will also change and improve. 

Once pretrained, the new model is saved and applied to text in the same way as the original XLM-R model. Importantly, this process can be extrapolated to other research questions, especially those in a very specialized fields for which the original XLM-R model does not have a strong set of pretraining text. We refer to the new pretrained model as XLM-R-Racismo in the paper --the full technical model name is `xlm-r-racismo-es-v2' and is now available for public use at \url{huggingface.co}. This is the model we apply below and the one to which we compare the performance of all models, including the original XLM-R.

\subsubsection{Fine-tuning our Models}

Before we fine-tune our models, we need to specify values for the hyperparameters. Following the recommendations by the authors of BERT and XLM-R and our own cross-validation tests, we use a batch size of 32, a learning rate of 2e-5, a maximum sequence length of 85, and 4 epochs.\footnote{Batch size refers to the amount of tokens that the algorithm will analyze simultaneously. For BERT, 16 or 32 are recommended for highest accuracy \citep{liu2019roberta}. Learning rate is the size of the steps used by the algorithm in each iteration towards a minimum of a loss function. The recommended learning rate for a BERT or RoBERTa model is $3e^{-5}$, however we modified the learning rate slightly to adapt it to our training set. The number of epochs is the number of times the model runs through the entirety of the data. For BERT and RoBERTa, the recommended number of epochs is between 2 and 4. The maximum sequence length is the number of token at which we truncate each observation (i.e., tweet). We do this to manage computer memory. We use 10-fold cross-validation for all models and report the averages.} Finally, we use the weighted Adam optimizer \citep{loshchilov2017decoupled}, which performs well for NLP data and for BERT and RoBERTa models --and their cross-lingual variants-- in particular.

We fine-tune two variants of our main XLM-R model: (1) the original XLM-R model\footnote{The version we use has been further pretrained for named entity recognition in Spanish.} and (2) our own XLM-R model pretrained specifically to detect racist text with Twitter data from Ecuador. To evaluate model performance, we use 10-times repeated cross-validation and report the usual metrics for NLP tasks, including accuracy, recall, precision, and F1-scores for each model. We are particularly interested in understanding how our model is misidentifying covertly and overtly racist language.\footnote{Depending on the application of this method, researchers might be more interested in the proportion of false negatives or false positives. This will depend on the ultimate goal of the project. We encourage researchers to pay particular attention to the possible effects of having more (or fewer) false positives or negatives on the conclusions obtained from estimations using these predictions.}  

We also train five separate models to serve as baselines, all of which are widely used in the social sciences and have been considered state-of-the-art at some point or another in the past decade \citep{grimmer2022text}. First are non-contextual word-embedding models such as GloVe and Word2Vec, which are usually applied via Convolutional or Recurrent Neural Network architectures (CNN or RNN). For this exercise, we use a CNN with Word2Vec and a Long-Short Term Memory RNN with GloVe embeddings. Bidirectional Long-Short Term Memory (Bi-LSTM) networks are a type of RNN that have proven particularly effective with NLP tasks due to their capacity to retain some contextual information from previous word sequences in the text. We are thus especially interested in comparing XLM-R's performance with that of a Bi-LSTM neural network using GloVe embeddings in the context of covert racism in Ecuador.

Additionally, we train two traditional machine-learning models to provide two non-neural network baselines: Support Vector Machine (SVM) and Logistic Regression (LR). For these two models, we tokenize the words using NLTK's Spanish word tokenizer, train and test the models, and use 10-fold cross-validation to evaluate model performance and results \citep{loper2002nltk}.\footnote{NLTK is a NLP library in Python. We use Python's Scikit-learn library to train and test these two models. We also use a TF-IDF vectorizer combined with the NLTK tokenizer with the SVM and LR models to improve performance.}

\subsubsection{Results and Validation}

\begin{table}[!b]
 \caption{Performance statistics for racist discourse classifiers (10-fold CV averages)\label{tab:performance}}
\scalebox{0.72}{\renewcommand{\arraystretch}{1.3}
\begin{tabular}{| l | c c c c | c c c c | c c c c | c c c c | c c c}\hline
\multicolumn{1}{|l|}{\textbf{Model}} & \multicolumn{4}{c|}{\textbf{Non-Racist}}  & \multicolumn{4}{c|}{\textbf{Covert Racism}} & \multicolumn{4}{c|}{\textbf{Overt Racism}} & \multicolumn{4}{c|}{\textbf{Macro-Average}}\\
 & Acc. & Prec. & Rec. & F1 &  Acc. & Prec. & Rec. & F1 &  Acc. & Prec. & Rec. & F1 &  Acc. & Prec. & Rec. & F1 \\  \hline
ML - Logistic Reg.  & 0.963 & 0.744 & 0.963 & 0.839 & 0.300 & 0.651 & 0.300 & 0.407 & 0.649 & 0.892 & 0.649 & 0.748 & 0.756 & 0.637 & 0.762 & 0.665 \\ 
ML - SVM & 0.925 & 0.808 & 0.925 & 0.862 & 0.451 & 0.611 & 0.451 & 0.514 & 0.746 & 0.849 & 0.746 & 0.789 & 0.783 & 0.708 & 0.756 & 0.722 \\ \hline
CNN - Word2Vec & 0.905 & 0.791 & 0.905 & 0.843 & 0.484 & 0.625 & 0.484 & 0.528 & 0.615 & 0.820 & 0.615 & 0.697 & 0.756 & 0.668 & 0.745 & 0.689  \\ 
Bi-LSTM - GloVe & 0.877 & 0.841 & 0.877 & 0.857 & 0.568 & 0.590 & 0.568 & 0.569 & 0.745 & 0.812 &  0.745 & 0.775 & 0.780 & 0.730 & 0.747 & 0.733 \\ \hline

XML-R & 0.865 & 0.890 & 0.865 & 0.876 & 0.759 & 0.712 & 0.759 & 0.733 & 0.761 & 0.788 & 0.761 & 0.771 & 0.818 & 0.795 & 0.797 & 0.794 \\

XML-R-Racismo & 0.904 & 0.927 & 0.904 & 0.915 & 0.817 & 0.797 & 0.817 & 0.805 & 0.834 & 0.808 & 0.834 & 0.818 & 0.851 & 0.803 & 0.844 & 0.846 \\ \hline
\end{tabular}}
\end{table}

The performance gains from the XLM-R models are considerable compared to the traditional machine learning models and CNN and Bi-LSTM using GloVe and Word2Vec embeddings. These results are in Table \ref{tab:performance}. Simple SVM and Logistic Regression models perform fairly well (perhaps better than expected) in identifying overtly racist and non-racist discourse, with accuracy scores over 90\% for the latter and over 60 and 70\%, respectively, for the former. The models do not fare as well with covert racism, which is much more difficult to detect without contextual understanding --both the logistic regression and SVM models score below 50\%. The CNN and Bi-LSTM models improve the classification for covert racism, as expected, given the more advanced word embeddings used and their greater capacity to understand text. The Bi-LSTM model detects covert racism more than 50\% of the time and performs well with overt racism, with an accuracy of 74.5\%.

The two XLM-R models in Table \ref{tab:performance} outperform the rest of the models. The original XLM-R model pre-trained by Facebook AI performs well despite not being pre-trained to detect racism in the Ecuadorian context. The XLM-R model identifies covert racism correctly 73.3\% (0.733) of the time in unseen data and identifies overt racism with 77.1\% accuracy (F1 score). The F1 score for non-racist discourse is also high at 0.876. Since this is an unbalanced panel, the F1 scores are a good reference point to validate accuracy scores and assess their differences across models. Overall, model F1 stands at 79.4\%, a high score considering the complexity of the task. Accuracy and F1 scores are consistent regarding overall model performance across all three categories.

Even more striking are the results of the XLM-R-Racismo model further pretrained on 1.7 million racist tweets from Ecuador (last row in Table \ref{tab:performance}). The model's accuracy scores are improved across the board compared to the original XLM-R model. The model shows strong improvement in covert racism, with an F1 score of 0.805---a 9.56\% improvement in performance when comprared to the original XLM-R model. Similarly, the model's F1 score for overt racism is 0.846, a 9.72\% gain. Regarding non-racist discourse, the model's F1 score increases more moderately from 0.876 to 0.915 for a 4.45\% gain. Overall, this is a substantial improvement in detecting covert and overt racism, especially considering we only used a relatively small corpus of 1.7 million tweets for pretraining, which pales compared to the 160GB of data used to pretrain the original XLM-R model. It is reasonable to think that larger amounts of task-specific pretraining data would further improve model performance during fine-tuning.

The results are particularly noteworthy given that similar work on detecting hate speech in Spanish text has achieved F1 scores for the hate-speech category of 72.7\% \citep{gertner2019mitre} and 75.5\% \citep{plaza2021comparing}, compared to our 81.8\% F1 score for the overt racism category. Both of these works used similar Transformers-based infrastructures based on Google AI's original models --specifically, they used mBERT (multilingual BERT) and BETO, a refined BERT model pretrained specifically with large amounts of text in Spanish. The performance gains of the XLM-R-Racismo are substantial compared to these two works (12.52\% and 8.34\% respectively).

Beyond the performance of our models, we are interested in conducting error analysis to identify the weaknesses of our best-performing set-ups. For the XLM-R-Racismo model, we check each mislabelled instance and compare instances that were commonly misclassified. Since we have a multi-class classification model, it is not sufficient to analyze the false positive (FP) or false negative (FN) rate, as it is not the same to mistakenly label a non-racist text as covertly racist and to label an overtly racist text as covertly racist. Thus, not only are we interested in identifying the FP and FN rate for each category, but more importantly, we are interested in identifying towards which category is the error skewing.

\begin{table}[!ht]
\centering{\begin{tabular}{l l l l l}\hline
& &\multicolumn{3}{c}{\textbf{True Category}} \\
 & & Non-racist & Covert racism & Overt racism \\ \hline
\multirow{3}{*}{\textbf{Predicted Category}} & Non-racist & \textbf{188} & 26 & 8 \\ 
& Covert     & 22  & \textbf{53} &  9 \\ 
& Overt     & 3  & 7 & \textbf{70}  \\ 
\end{tabular}

}
 \caption{Confusion matrix for the pretrained XLM-R model\label{tab:confusion}}
\end{table}

Table \ref{tab:confusion} shows the confusion matrix for the XLM-R-Racismo model. Our model is accurate at predicting overtly racist and non-racist text. In the model, text that is misidentified as overtly racist is more likely to be covertly racist (8.8\%) rather than non-racist (3.8\%). This is a positive result, as covertly and overtly racist text is, above all, racist text (just different manifestations of the phenomenon). As is evident from all models, covert racism is more difficult to identify. XLM-R-Racismo is more likely to misidentify covertly racist text as non-racist (26.2\%) than overtly racist (10.7\%), which is consistent with our expectations. More important, however, is the relatively low FP rate for non-racist text. The XLM-R-Racismo model misidentifies (15.3\%) of non-racist text was classified as either covertly or overtly racist.

Having shown the advantages and limitations of our approach, we use our trained model to predict racist discourse in a corpus of 143,733 unique tweets collected during the 2019 ind\'igena strike in Ecuador. Following the same rules used to label our training data set, we manually code a sample from the predicted data. We are interested in the extent to which our model is able to scale up to new text. We present the performance statistics in Table \ref{tab:confusion_oos}. Compared to the results from Table \ref{tab:performance}, this out-of-sample performance is similar to the ones obtained with the training data.

\begin{table}[!ht]
\centering{\begin{tabular}{l l l l l}\hline
& &\multicolumn{3}{c}{\textbf{True Category}} \\
 & & Non-racist & Covert racism & Overt racism \\ \hline
\multirow{3}{*}{\textbf{Predicted Category}} & Non-racist & \textbf{214} & 1 & 0 \\ 
& Covert     & 40  & \textbf{92} &  1 \\ 
& Overt     & 21  & 10 & \textbf{96}  \\ 
\end{tabular}

}
 \caption{Confusion matrix for XLM-R-Racismo predictions and hand-coded out-of-sample data \label{tab:confusion_oos}}
\end{table}

\section{Covert and Overt Racism in Ecuadorian Twitter} 

Our exploratory hypotheses suggest that covert racism will be more prevalent than overt racism. Since overt racism is socially costly, public figures will avoid producing overt racism at a greater rate than non-public figures. Furthermore, challenges to the status of the blanco-mestizo population in Ecuador will increase the prevalence of racist discourse.  overt racism will increase at a higher rate than covert racism. To test our hypotheses and to showcase the usefulness of our approach, we use our trained model to predict racist discourse on the two corpora described in the Data section: Twitter data during the 2019 Indígena Protest in Ecuador (Protest Data), and Twitter data mentioning the Indígena community between 2018 and 2022 (Indígena Data). In addition to the text, we obtained information on the user that created the tweet, the user(s) that retweeted the message, the time when each message was tweeted, and the status of the sender (e.g., verified). We present results and analyses for each corpus separately to estimate variations across time and source. In Table \ref{tab:allstat}, we show the distribution of non-racist, covertly racist, and overtly racist tweets in both corpora. 

\begin{table}[!ht]
\centering \caption{Prevalence of Different Forms of Racist Tweets \label{tab:allstat}}
\begin{tabular}{l c c c c}\hline
 & \textbf{No Racism} & \textbf{Covert Racism} & \textbf{Overt Racism} & \textbf{Total} \\ \hline
\textit{Corpus Protest Data} & 159,314 & 7,312 & 1,436 & 168,062\\ 
\textit{Corpus Indigena Data} & 144,520  & 14,350 & 3,315 & 153,769\\ 
\end{tabular}
\end{table}  

\begin{figure}[!ht]
    \centering
    \subfloat[\centering Primary connected network]{{\includegraphics[width=5cm]{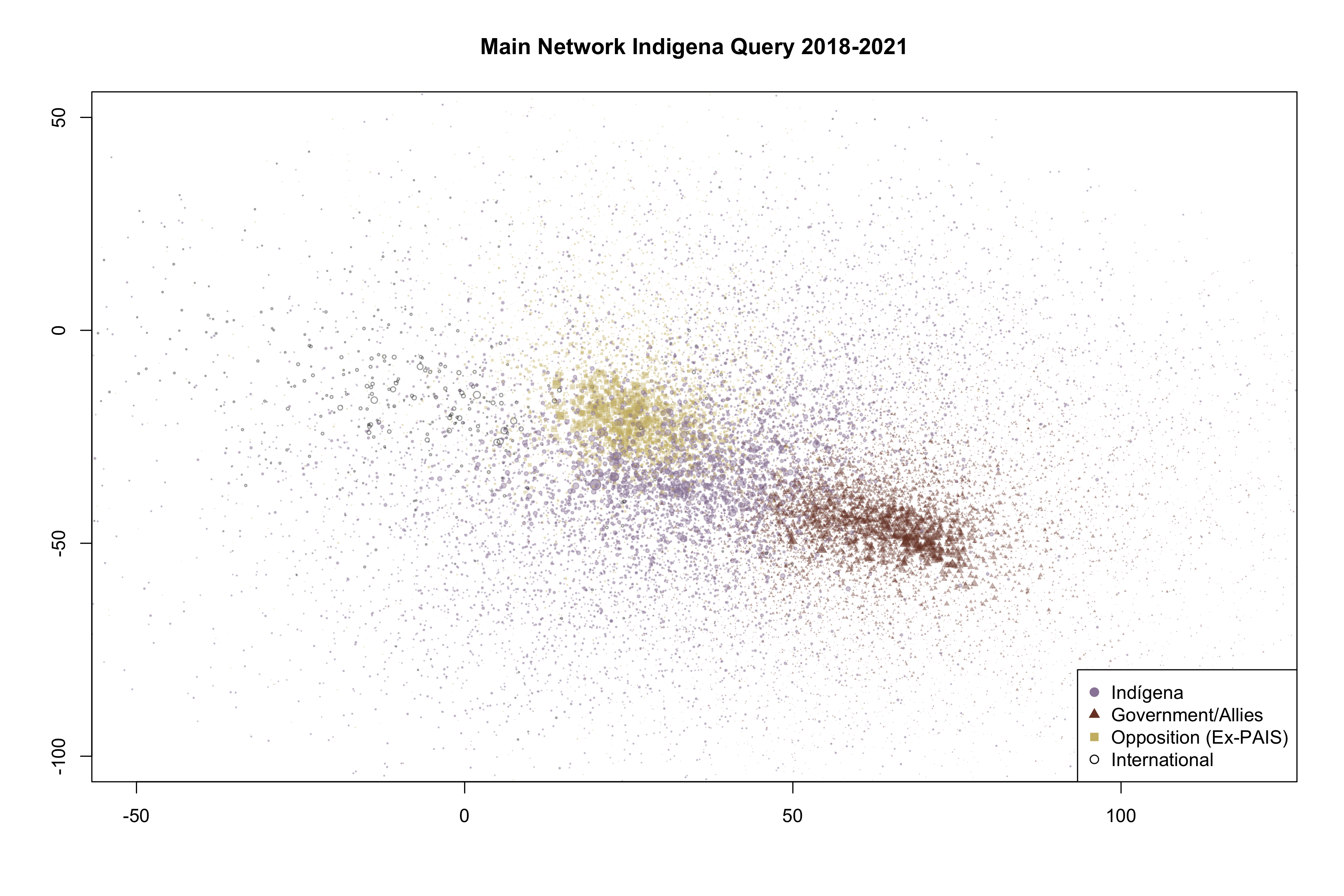}}}%
    \qquad
    \subfloat[\centering Nodes engaging with covertly racist content]{{\includegraphics[width=5cm]{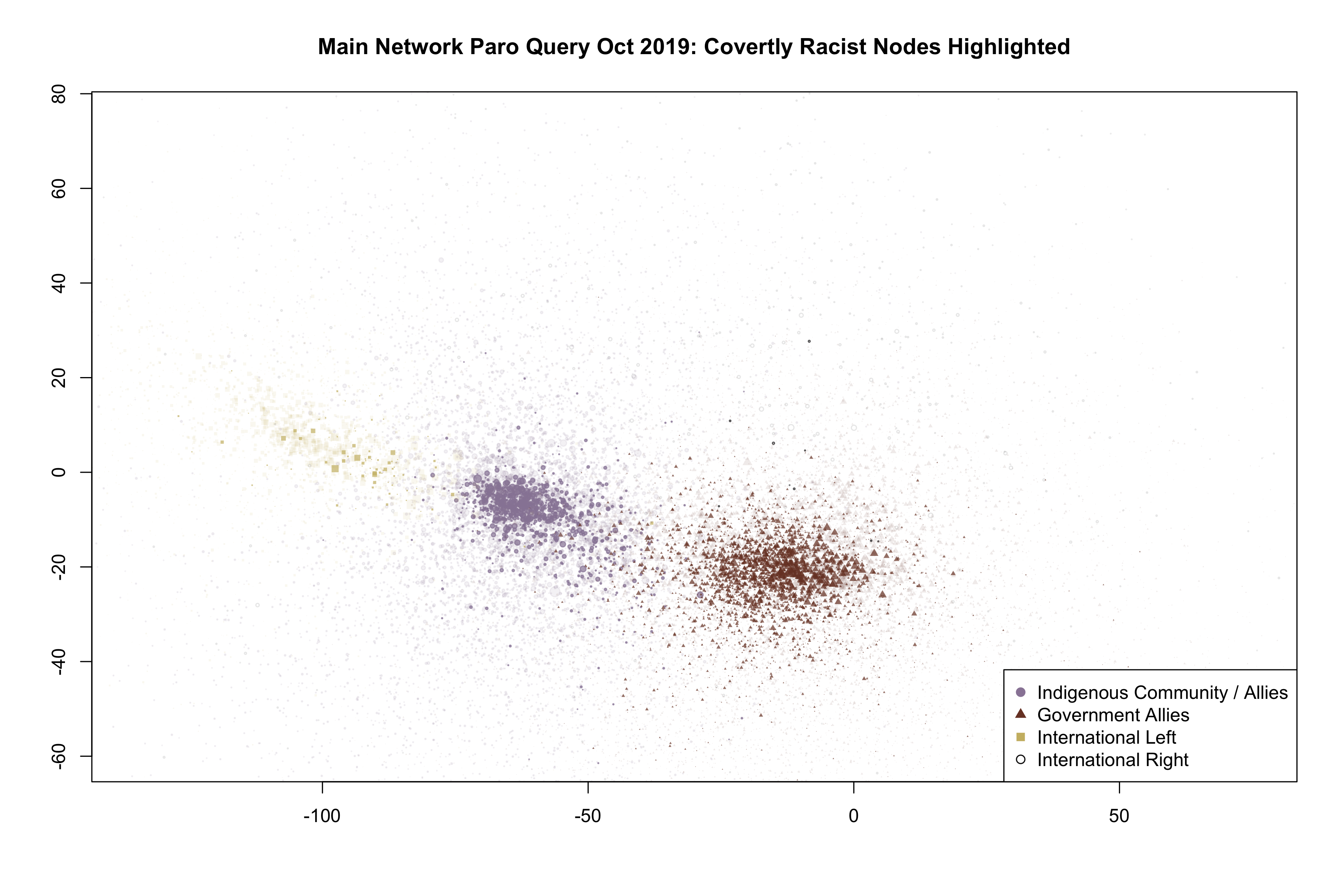}}}%
    \qquad
    \subfloat[\centering Nodes engaging with overtly racist content]{{\includegraphics[width=5cm]{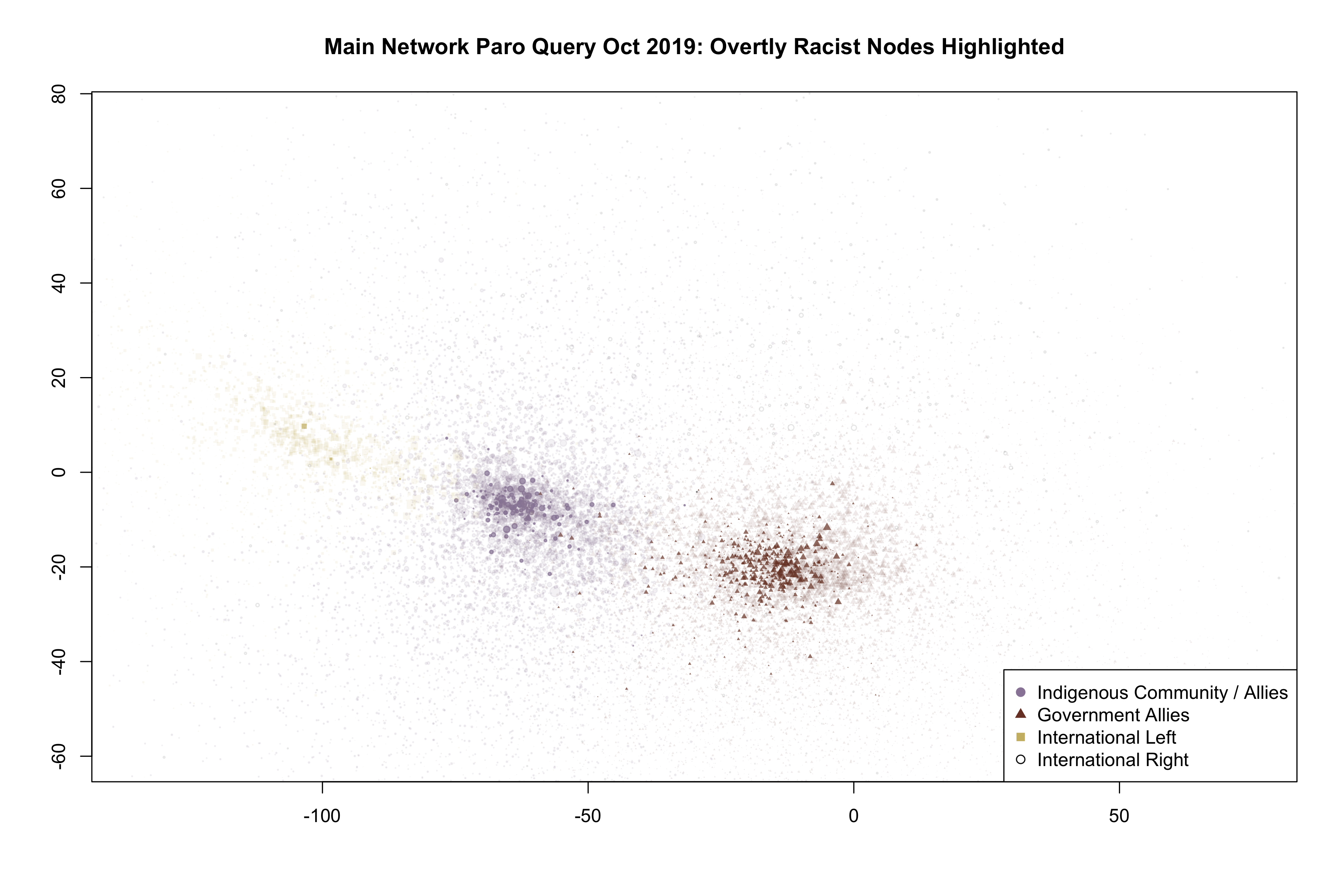}}}
    \caption{Left panel: Primary connected network during the Ecuadorian protests, between October 1 and October 24 2019.Center panel: Nodes engaging with covertly racist tweets during the Ecuadorian protests. Right panel: Nodes engaging with overtly racist tweets during the Ecuadorian protests.Brown triangles describe pro-government users. Purple circles describe users aligned with the ind\'igena community. Yellow squares describe users aligned with the pro-Correa community. Hollow circle describe users from the ``International Right.''}
\label{fig:net_racism}
\end{figure}

We identify communities --clusters of nodes where the same information (tweets) is shared-- via a random-walk community detection algorithm \citep{pons2005computing}.\footnote{A random-walk community detection algorithm is based on the idea that a random walk (walking randomly from one connected node to another) will tend to stay within a community instead of jumping from across communities \citep{pons2005computing}. We provide more details on how we create the networks and the choice of our community detection algorithm in Appendix F.} One characteristic of Twitter networks is their consistency across time \citep{calvo2020fake}. Thus, it is not surprising that the random-walk community detection algorithm identified the same three primary communities in both datasets: a pro-government network, which includes 41,493 nodes in the Protest Data and 26,036 nodes in the Indígena Data; an indígena community network of 30,244 nodes in the Protest Data and 26,110 in the Indígena Data; and a pro-Correa community network of 15,635 nodes in the Protest Data and 11,325 in the Indígena Data.\footnote{While not political allies to the indígena community, the pro-Correa community was in opposition to the government.} The communities identified align with the main political factions during the protests \citep{diazelites}. A visual representation of the communities for both the Protest dataset is shown in the left panel of Figure \ref{fig:net_racism}. We limit our analysis to the two main communities: the pro-government community and the pro-indígena community. 


From $H1_a$, we expect a higher prevalence of covert racism than overt racism. In both corpora, the prevalence of predicted covert racism exceeds that of overt racism (see Table \ref{tab:allstat}). We can also visualize nodes that engage with racist tweets, either by tweeting or by retweeting racist content. In the center panel of Figure \ref{fig:net_racism}, we highlighted nodes that engaged with covertly racist tweets, and in the right panel, we highlighted nodes that engaged with overtly racist tweets. Notice how highly engaged nodes--nodes closer to the center of each community--are more likely to participate in racist discourse.

However, precisely because overtly racist language is socially punishable, more so than covert racism, we also expect public figures to avoid producing and reproducing racism in its most overt form ($H1_b$). To detect public figures, we identify `verified' accounts.\footnote{During the period of data collection, verified accounts had a thorough vetting process in Twitter. Verified status was granted only to public figures, government representatives, news organizations, and journalists.} We find $H1_b$ to be true for our model. Verified accounts--political figures and media accounts--do not produce or reproduce overtly racist content. The change of exposure and the cost public figures incur when engaging with racist content surpasses that of less public users. Most verified users are news outlets and government officials who often have staff managing their social media accounts and have more to lose if they were to engage in overt forms of racism publicly.

\begin{figure}[!ht]
\centering
\includegraphics[width=16cm]{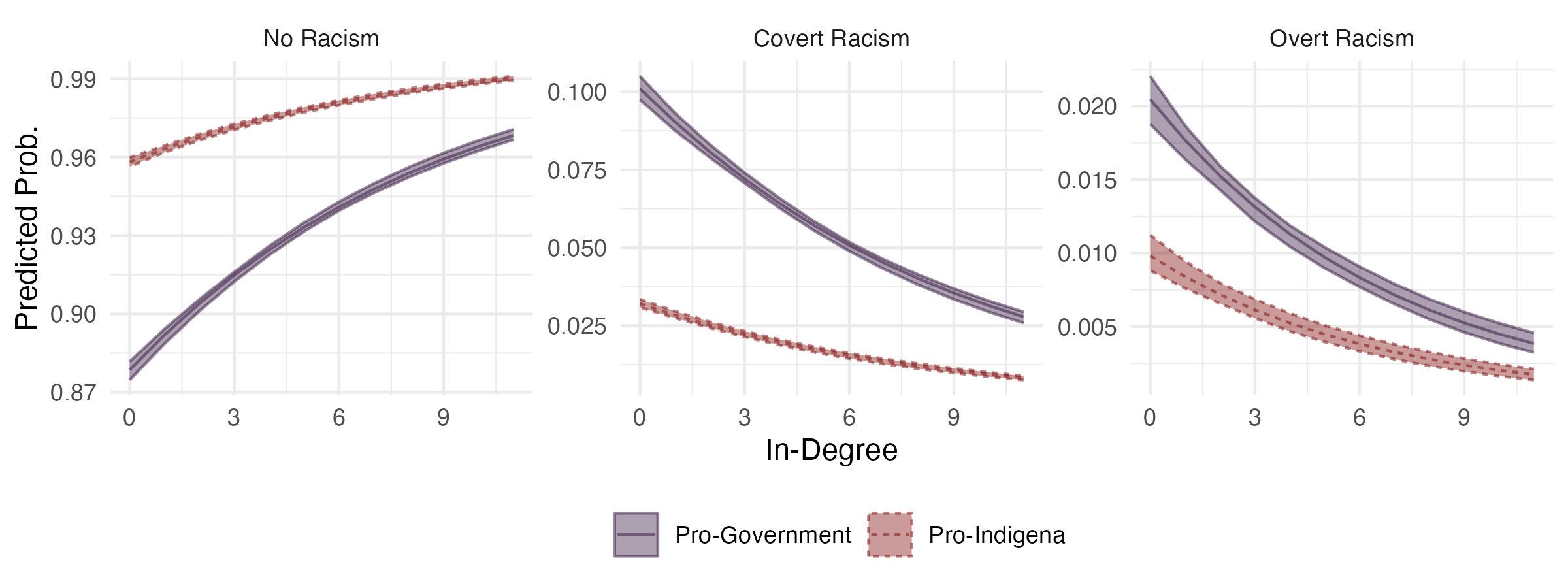}
\caption{Predicted probabilities of tweeting covertly, overtly, or none racist content during the Ecuadorian protests (2019). We find a similar result using the data from the network mentioning the indígena community (2018-2022).}
\label{fig:pred_prob_h1b}
\end{figure}

The opposite is also true: less influential users (i.e., users with low in-degree) and users in the periphery of the network topography produce most of the overtly racist content. To formally show this relationship, we regress a multinomial logistical model with a categorical dependent variable for whether the tweet is non-racist, covertly racist, or overtly racist, against the (log) in-degree of the user tweeting, as well as the community to where they belong. We plot the predicted probabilities in Figure \ref{fig:pred_prob_h1b} for ease of interpretation. The results suggest that, as in-degree increases, the probability of tweeting racist content decreases. For example, going from the first quartile of the in-degree distribution to the third quartile reduced the probability of producing racist content by half. Also noteworthy is the difference in the prevalence of covertly and overtly racist content between pro-government and pro-indígena communities. For example, of all overtly racist content in the Protest Data (6,446 retweets), users from the pro-government retweeted 75.4\%, while users from the pro-ind\'igena community retweeted 20.6\%. Most of the overtly racist content from the ind\'igena community was either wrongly categorized nodes (58.5\%) or wrongly predicted tweets (35.5\%).\footnote{Most of the wrongly predicted tweets in this corpus were texts where it was not clear to whom a particular insult was directed. While insulting somebody is not a racist act, doing so by priming their identity is. For example, in the tweet: `CONAIE [sic] attacking the wrong person. There are incompetent police officers, just as idiotic as the thugs.' it is not clear if the user calling `idiotic' and `thug' the CONAIE, which is a reference to the indígena identity. Our model in these cases assumes that the user is, in fact, referring to the indígena identity.}  

Covert racism is spread across all communities. While covert racism is still more prevalent within the pro-government community, we see instances of this type of discourse in the pro-Correa and pro-ind\'igena communities. This is to be expected: some covert forms of racism are embedded in apparently benign intentions (see Section 4). Users from the pro-government community retweeted 77.2\% of all covertly racist content (37,025 retweets), while users from the ind\'igena community retweeted 18.8\% (9,076 retweets). Covert forms of racism from supporters of the ind\'igena community often perpetuate the idea of ind\'igenas as a cursed race, a people in need of protection (i.e., \textit{infantilization}), accused indígena leaders that did not support the strike of not being ``real'' indígenas (i.e., \textit{denying their identity}), or questioned whether those leaders involved in the strike were ``real'' indígenas.   For example, a user supporting the strike tweets:  ``Yaku Pérez [an indígena leader] is a fake indígena, trying to [take advantage of the situation]. CONAIE cannot allow for this treason.'' 

While users at the center of the network topography are less likely than less prominent users to tweet covertly racist content, they do so at a higher rate than overtly racist content. The lower cost of these discursive forms allows more prominent and more public users --users whose reputation would suffer from engaging with overtly racist content-- to produce and reproduce it. This includes members of the media (e.g., online political magazine @4pelagatos), politicians (e.g., former-president @mashirafael and then-president @lenin), and political commentators (e.g., pro-government journalist @CarlosVeraReal). For example, a common characterization of the indígena community during the protests was as ``terrorists'' or ``criminals'' (one of the characteristics of ``hygienic racism''). Then-President Lenin Moreno tweeted: ``The indígena leadership, supportive of Democracy and the Rule of Law for all Ecuadorians, must stir away from these false leaders, akin to indigenous terrorism and paramilitarism.''

\begin{figure}[!ht]
    \centering
    \includegraphics[width=16cm]{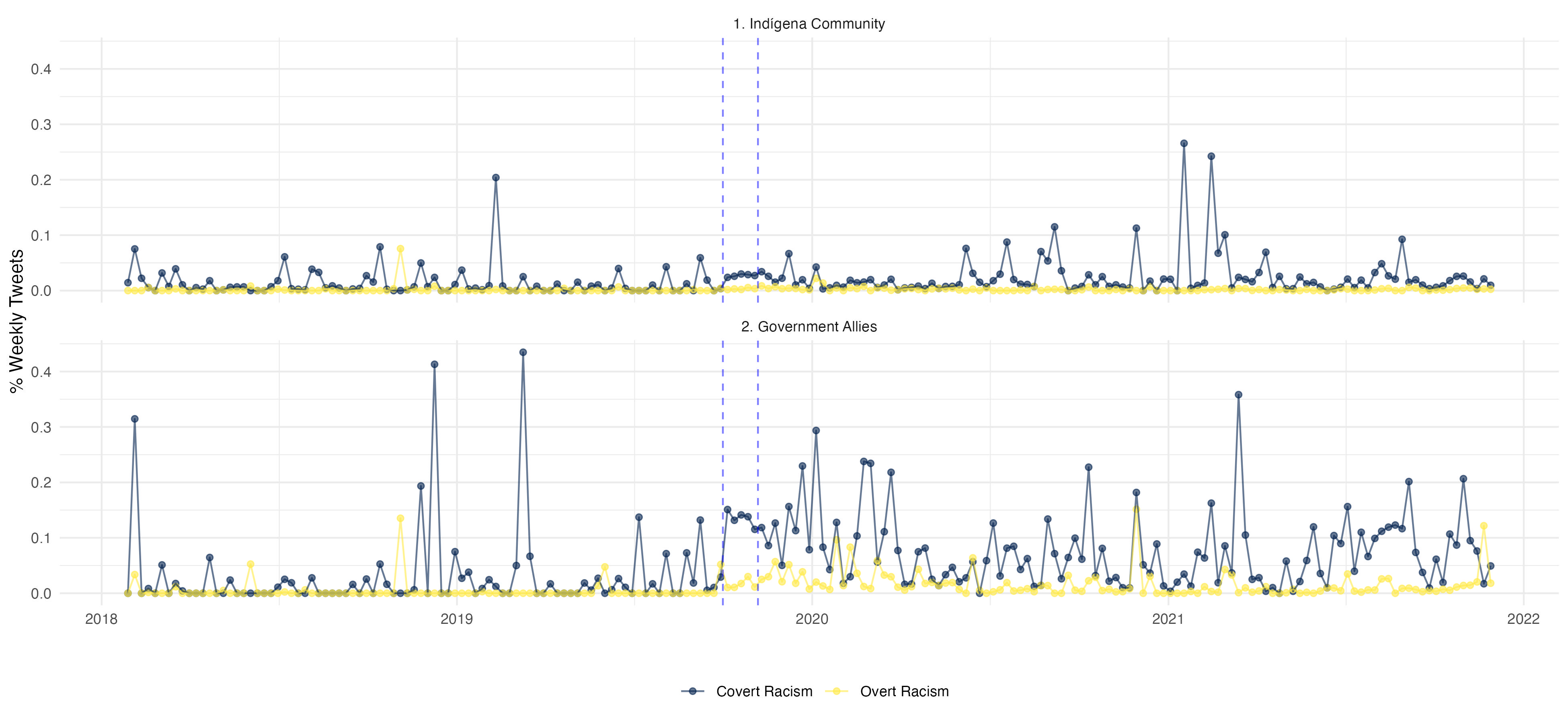}
\caption{Percentage of weekly covertly and overtly racist tweets reproduced in the Indígena Data, between 2018 and 2021.
}
\label{fig:racism_timeline}
\end{figure}

To support our second set of hypotheses, we look at the prevalence of covert and overt racist tweets across time. We are particularly interested in the change in the prevalence rate of covertly and overtly racist tweets during the indígena strike of 2019. To analyze the long trend, Figure \ref{fig:racism_timeline} shows the weekly percentage of covertly and overtly racist tweets reproduced between 2018 and 2021. The highlighted space marks October 2019, the month of the indígena strike, where there is a marked increase in the number of racist tweets (both covert and overt). As suggested in $H2_a$, racist discourse will be more prevalent when the dominant group's status is challenged. We also find that covertly racist discourse is more likely to be reproduced during `normal times' than overtly racist tweets. This speaks to the normalized nature of certain forms of racism, particularly covert ones. 

We can further analyze how users engage with racist discourse when the status of the white-mestizo population is challenged. To measure engagement, we follow Aruguete and Calvo \citeyear{aruguete2018time} and estimate time-to-retweet, the difference between the time a user posted a tweet and the time a second user retweeted it. Time-to-retweet serves as a proxy to acceptance (less time-to-retweet, more acceptance of the content) and cognitive congruence (less time-to-retweet, more cognitive congruence) \citep{aruguete2018time}. We can analyze the effects of challenges to the status of the white-mestizo population by examining the moment when Jaime Vargas, CONAIE's president, called the police and military forces to disobey government orders and join the protest.\footnote{On October 10, the indígenas assembled in a large forum in Quito where eight police officers, including a colonel, were detained for more than two hours, as well as various journalists and their crews. Jaime Vargas, the president of the CONAIE, spoke to the crowd on live television with the detained police officers in the backdrop. At 10:30 AM local time, Vargas asks the detained colonel, the Police force, and the military to join the protests and disobey the orders of the government.} We can estimate the effect of Varga's actions on users' engagement with racist content as we have data on the exact time of the call.  We follow C \cite{calvo2023winning} and \cite{vallejo2023rage} and employ an interrupted time series analysis, a variety of regression discontinuity designs (RDD) in which the running variable is time \cite{morgan2015counterfactuals}. We estimate average changes in time-to-retweet at the cut-off point. The key assumption of this approach is that any immediate change in the time-to-retweet can be attributed to the events and not to any other factor affecting time-to-retweet that also changed systematically at the same time. The granularity and abundance of Twitter data, as well as the public nature of this event, allows us to isolate the change in time-to-retweet to the specific second the event took place. We can thus make the plausible assumption that other omitted variables (e.g., other events not attacking the status of the in-group) are not also changing suddenly at the time of the events. To further isolate the event's effect, we estimate the models within a four-hour window around the cut-off (i.e., the time of the event).\footnote{We limit our analysis to the Protest Data.}

\begin{figure}[!ht]
\centering
\includegraphics[width=16cm]{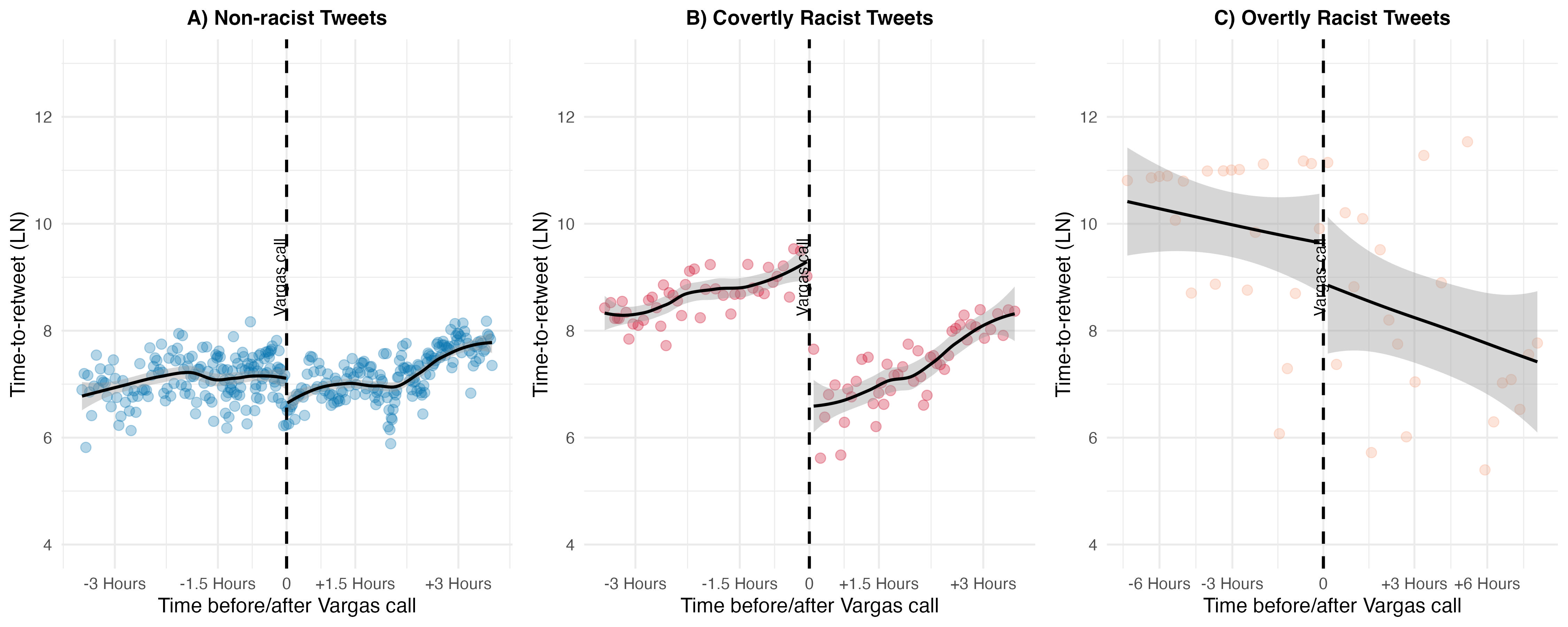}
\caption{Changes in Time-to-Retweet during Vargas' call (pro-government community only).}
\label{fig:rdd_vargas}
\end{figure}

The dependent variable is time-to-retweet, and the cut-off time of the discontinuity is centered on October 10, 2019, at 10:30 AM local time, when Jaime Vargas, the president of the CONAIE, asks the Police force and the military to join the protests and disobey the orders of the government. We focus solely on the pro-Government community,\footnote{In Appendix G we present results for the pro-Indígena community. However, we find no significant changes in time-to-retweet for any of the categories.} and estimate different models for non-racist, covertly racist, and overtly racist tweets. The treatment effect (i.e., the event) on racist tweets is negative--time-to-retweet is reduced--and statistically significant at the 95\% level. Figure \ref{fig:rdd_vargas} plots the regression discontinuity results for the event to better illustrate the effect. The vertical axis reports time-to-retweet, where lower values mean less time-to-retweet and less user latency. The horizontal axis ranges from six hours before and after the event. We use a LOESS smoother to fit the underlying regression function separately before and after the event. The discontinuity shows no latency change for users sharing non-racist content at the time of Vargas' speech. However, there is a statistically and substantively significant difference in latency for users sharing covertly racist messages (p $\leq$ 0.01). Immediately after Vargas' call, covertly racist content increases engagement and reduces latency among pro-government users by 27.8\%. Furthermore, before the event we find that on average users tweet at a slower rate covertly racist content than non-racist content. The event causes the latency for covertly racist tweets to be the same as non-racist tweets. While the direction of the effect on overtly racist tweets is as expected, the relation is not statistically significant at conventional levels. However, note that there are only 50 effective observations around the cut-off, so the results must be interpreted cautiously. Overall, the reaction to the perceived threat to the status of white-mestizos points to pro-government users accepting covertly racist frames and increasing the speed of their diffusion. The same is not true for overtly racist tweets.

\section{Limitations}

We identify five limitations to our approach. First, racist text is a rare event in `naturally-occurring' corpora, which is likely to produce unbalanced datasets for training. Second, the model more often mislabelled racist discourse as non-racist than non-racist discourse as racist (more false negatives than false positives). Third, there may be contextual clues in racist text that NLP models may miss. Fourth, racist discourse evolves over time, and researchers need to incorporate these discursive changes in their approach. Fifth, our approach requires both computational power and skills for implementation. We have made the code accessible and provided guidance on how to use these models, but computational barriers may remain. We provide a detailed discussion of each of these limitations in Appendix H.


\section{Conclusion} 

This article provides a methodological approach to classify different forms of racist discourse in text that combines a theoretical understanding of the context with state-of-the-art machine learning approaches. Given the contextual nature of racism, we highlight the importance of creating coding rules that consider the origin and manifestations of racism in the place and time of study. Having theoretically grounded rules to categorize different forms of discursive racism is the first step to identifying them in large corpora.

The second step uses a Transformers-based deep-learning approach to text classification. The Transformers architecture has revolutionized NLP tasks, partly for its ability to better understand context compared to other state-of-the-art deep learning approaches, like CNN and RNN models. Transformers-based models such as XLM-RoBERTa can be further pre-trained with specialized text that improves classification performance. Adequately trained models can accurately classify text that requires a nuanced understanding of context, such as racist discourse. We apply this process to identify covert and overt racism in a corpus of 3M+ tweets relating to the indígena community in Ecuador between 2018 and 2021. Our main model XLM-R-Racismo, an XLM-RoBERTa model further pre-trained using 1.7 million tweets, outperforms other machine-learning models in detecting overt and covert racism in text. 

The results from our analysis highlight not only the usefulness of our method but also the implications on how we understand the role of racist content in modern settings. We found the expected patterns of overtly racist content in social media: more central nodes (i.e., more important users) are less likely than more peripheral nodes to tweet racist messages. The effect was weaker on covertly racist tweets. The consequences of the characteristics of racist language in current-day Ecuador (i.e., language that is socially punished) are replicated in social media. In this context, we also found that users react faster to racist tweets when there is a challenge to their in-group status. Racist language seems to serve a rallying purpose, a mechanism to reinforce or remind the out-group of their position within the racist structure.

Finally, while the primary goal of this paper is to provide a step-by-step approach to classifying racist discourse in large corpora, we believe our contribution can be extended to other areas. The ability of Transformers-based approaches to understanding language in context makes them an ideal tool to classify text that requires a contextual understanding. The identification of covert and overt racism serves as an example of the possibilities. Researchers can exploit the approach presented here to answer other substantive questions in the discipline, using the vast corpora available in new and creative ways.

\setstretch{1}
\bibliography{refs.bib}
\bibliographystyle{apsr}

\end{document}